\documentclass[journal]{IEEEtran}

\ifCLASSINFOpdf
\else
   \usepackage[dvips]{graphicx}
\fi
\usepackage{url}

\usepackage{graphicx}
\usepackage{cite}
\usepackage{amsmath,amssymb,amsfonts}
\usepackage{algorithmic, algorithm}
\usepackage{textcomp}
\usepackage{xcolor}
\usepackage{multirow}
\usepackage{booktabs}
\usepackage{bbding}
\usepackage{bm}
\usepackage[colorlinks,
linkcolor=black,
anchorcolor=blue,
citecolor=black
]{hyperref}

\begin{document}

\title{Joint Selective State Space Model and Detrending for Robust Time Series Anomaly Detection}

\author{Junqi Chen, Xu Tan, Sylwan Rahardja, \IEEEmembership{Graduate Student Member, IEEE},\\ Jiawei Yang, \IEEEmembership{Member, IEEE}, and Susanto Rahardja, \IEEEmembership{Fellow, IEEE}
\thanks{Junqi Chen, Xu Tan and Susanto Rahardja are with the School of Marine Science and Technology, Northwestern Polytechnical University, Xi’an, 710072, China. (e-mail: jqchen@ieee.org; xutan@ieee.org; susantorahardja@ieee.org).}
\thanks{Sylwan Rahardja is with the School of Computing, University of Eastern Finland, FI-80101 Joensuu, Finland. (e-mail: sylwanrahardja@ieee.org).}
\thanks{Jiawei Yang is with the Department of Computing University of Turku, 20014, Turku, Finland (e-mail: jiaweiyang@ieee.org).}
\thanks{Susanto Rahardja is also with the Engineering Cluster, Singapore Institute of Technology, 10 Dover Drive, Singapore 138683.}
\thanks{(Corresponding author: Susanto Rahardja.)}
}

\markboth{Journal of \LaTeX\ Class Files, Vol. 14, No. 8, August 2015}
{Shell \MakeLowercase{\textit{et al.}}: Bare Demo of IEEEtran.cls for IEEE Journals}
\maketitle

\begin{abstract}
Deep learning-based sequence models are extensively employed in Time Series Anomaly Detection (TSAD) tasks due to their effective sequential modeling capabilities. However, the ability of TSAD is limited by two key challenges: (i) the ability to model long-range dependency and (ii) the generalization issue in the presence of non-stationary data. To tackle these challenges, an anomaly detector that leverages the selective state space model known for its proficiency in capturing long-term dependencies across various domains is proposed. Additionally, a multi-stage detrending mechanism is introduced to mitigate the prominent trend component in non-stationary data to address the generalization issue. Extensive experiments conducted on real-world public datasets demonstrate that the proposed methods surpass all $12$ compared baseline methods.  
\end{abstract}


\IEEEpeerreviewmaketitle

\section{Introduction}
As the volume of data generated continues to grow exponentially, Time Series Anomaly Detection (TSAD) has garnered significant attention due to its increasing demand in various real-world applications such as intrusion detection, disaster warning, and medical diagnosis \cite{darban2022deep, yang2020classification, yang2022mipo, larosa2022separating, wang2021main}. TSAD aims to identify irregular points or subsequences collectively referred to as \textit{anomalies}. Typically,
samples with high reconstruction errors of deep neural networks (DNNs) trained exclusively on normal data are detected as anomalies.
Given the temporal correlations inherent in time series \cite{lai2021revisiting}, DNN-based sequence models are considered the most suitable approach for TSAD tasks \cite{cheng2023learning, hundman2018detecting,su2019robust, du2021gan,xu2021anomaly,tuli2022tranad,yang2023dcdetector,wang2024drift}. However, two key challenges remain in these sequence model-based TSAD: (i) the ability to model long-range dependencies and (ii) the generalization issue for non-stationary data.

{\color{black} Normal behavior in time-series data typically involves long-term dependencies\cite{darban2022deep}. To capture these dependencies and enhance the modeling of normal behavior, various sequence models including Temporal Convolutional Networks (TCNs) \cite{he2019temporal}, Recurrent Neural Networks (RNNs) \cite{hundman2018detecting, su2019robust} and Transformers \cite{xu2021anomaly, tuli2022tranad} have been explored for TSAD.} 
However, existing methods still encounter difficulties due to their intrinsic characteristics, such as limited context window, high memory costs, or unstable gradient flow. 
Recent research on the selective state space model (S6) \cite{gu2023mamba} has addressed the shortcomings of the aforementioned sequence models and demonstrated its excellent long-term modeling capabilities in other domains. Moreover, the selective nature of the S6 model allows it to discard abnormal information and generate reliable reconstructed output for TSAD. Despite offering a potential solution to model long-range dependencies, there has been no research exploring its application in TSAD.

The generalization issue for non-stationary data stems from trends in non-stationary data. 
The erratic distribution of trends can result in significant fluctuations in data magnitude, leading to erroneously high reconstruction errors in regions with previously unseen trend patterns in the training set, ultimately causing false alarms \cite{wang2024drift}.
Traditional time series decomposition methods such as Seasonal-Trend decomposition using LOESS (STL) \cite{cleveland1990stl} and Hodrick–Prescott (HP) trend filter \cite{hodrick1997postwar} are frequently used to mitigate the impact of these trends. However, achieving optimal decomposed results may not yield the best detection performance. To address this, several approaches have tried to integrate decomposition methods into detectors for joint optimization. Nevertheless, most of them rely on Moving Average (MA) with a fixed kernel size \cite{wu2021autoformer, zhou2022fedformer, wang2024drift}, lacking the flexibility for broader scenarios. {\color{black}Other methods rely on DNN models for detrending \cite{zhang2024unravel,qin2022decomposed}. Though they allow end-to-end optimization, their dependence on training data reintroduces generalization issues.}

To tackle the challenges outlined above, an innovative detector constructed using the S6 model and integrated with a multi-stage detrending mechanism is proposed. 
The contributions of this paper is as follows: (i) Introduce a novel detector that utilizes the S6 model to effectively model long-range dependencies in TSAD. (ii) Propose a multi-stage detrending mechanism that can generate reliable decomposed results. (iii) Evaluate the proposed method on three benchmark datasets and demonstrate the superiority of the proposed method over recent State-of-the-art (SOTA) methods.

\section{Related Work}
\begin{figure}[ht]
	\vspace{-1.0em}
	\centering
	\includegraphics[width=8.5cm]{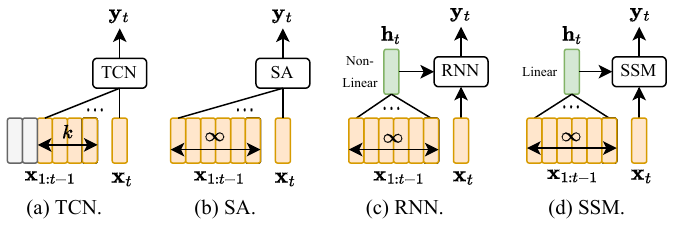}
	\vspace{-1.0em}
	\caption{Illustration of different sequence models gathering context information, where $k$ denotes the kernel size, $t$ denotes the time-step, $\bm{x}_{1:t-1}$ represents the context information, $\bm{x}_t$ and $\bm{y}_t$ represent the input and output in current time-step, respectively. (a) TCN's context was constrained by the kernel size $k$. (b) SA in the Transformer could capture long-term context but lacked information compression. (c-d) RNN and SSM could compress context information by a finite state.}
	\label{fig:context}
	\vspace{-2.0em}
\end{figure}

Sequence models such as TCN, RNN, and Transformer \cite{bai2018empirical, hochreiter1997long, vaswani2017attention} had greatly improved TSAD tasks. However, these methods struggled with long-range dependency modeling, limiting detection performance. Fig. \ref{fig:context} illustrates how different sequence models gather context information. As shown in Fig. \ref{fig:context}(a), TCN's context was constrained by the kernel size $k$, which made it unsuitable for capturing long-term dependencies. Conversely, the Transformer could capture long context using the Self-Attention (SA) mechanism as shown in Fig. \ref{fig:context}(b). However, SA was memory and time-intensive during training and inference because it did not compress context information, which made it impractical for modeling long dependencies. Although RNN could efficiently compress context information using a finite state shown in Fig. \ref{fig:context}(c), it suffered from unstable gradient issues with long sequences \cite{pascanu2013difficulty}. The State Space Model (SSM), similar to RNN, resolves gradient issues by employing solely linear transformations, as shown in Fig. \ref{fig:context}(d). 

The Structured State Space Sequence model (S4) \cite{gu2021efficiently} was a well-known SSM that combines the benefits of RNN with the stability of linear transformations. This allows S4 to compress context states into a finite size while ensuring stable gradients, making it suitable for modeling long-range dependencies.
However, its parameters remain constant through time, forming a Linear Time Invariance (LTI) model, limiting its effectiveness in TSAD tasks.
To remove the LTI constraint, Gu et al. \cite{gu2023mamba} proposed a selective SSM called S6 by introducing input-dependent parameters. S6 exhibited superior performance in modeling long-range dependencies in other domains, including natural language processing \cite{gu2023mamba}, computer vision \cite{liu2024vmamba}, and speech \cite{jiang2024dual}. 
Additionally, its selection mechanism has the potential to enhance TSAD performance by adaptively selecting the appropriate context information for producing reliable anomaly scores. Despite its effectiveness across various domains, its application in TSAD remains to be unexplored.

\section{Preliminaries}

\subsection{Problem Statement}
Given a multivariate time series $\bm{x} \in \mathbb{R}^{T \times D}$ with length $T$ for training, where each observation $\bm{x}_t \in \mathbb{R}^D $ contains $D$ features. TSAD aims to predict anomalous labels $\bm{y} = [y_1, \cdots, y_{\hat{T}}]^T$ for unseen test time series $\hat{\bm{x}}$ of length $\hat{T}$, where $y_t \in \{0, 1\}$. 
This prediction is based on anomaly scores $\bm{a} = [a_1, \cdots, a_{\hat{T}}]^T$ generated by the detector and a predefined threshold $a_{\rm th}$, where $y_t = 1$ if $a_t > a_{\rm th}$, otherwise $0$.

\subsection{S6 Model}
A typical SSM model maps a univariate sequence $x_t \in \mathbb{R}$ to the output $y_{t} \in \mathbb{R}$ through a hidden state $\bm{h}_{t} \in \mathbb{R}^N$ with four parameters $\Delta\in \mathbb{R}$, $\bm{A}, \bm{B}, \bm{C} \in \mathbb{R}^N$, where $t$ denotes the $t$-th time-step and $N$ is the number of states. 

To remove the LTI constraint, three of the above parameters in S6 model were adjusted to vary with time, namely $\bm{\Delta}\in \mathbb{R}^{T}$, $\bm{B}, \bm{C} \in \mathbb{R}^{T\times N}$, where $T$ denotes the number of time-steps.  The operation of the S6 model could be represented by a general formula applicable to multivariate time series data $\bm{x}, \bm{y} \in \mathbb{R}^{T\times D}$:
\begin{equation}
	\begin{aligned}
		\bm{h}_{t, d} &= \overline{\bm{A}}_{t, d} \odot \bm{h}_{t-1, d} + \overline{\bm{B}}_{t, d} \cdot {x}_{t, d}, \\
		{y}_{t, d} &= \bm{C}^T_{t} \cdot \bm{h}_{t, d},
	\end{aligned}  \label{eq:s6}
\end{equation}
where $d$ denotes the $d$-th feature, $\odot$ represents the Hadamard product, $\overline{\bm{A}}_{t, d} = f_A({\Delta}_{t, d}, \bm{A}_d)$ and $\overline{\bm{B}}_{t, d} = f_B({\Delta}_{t, d}, \bm{B}_{t})$. Here, $f_A(\cdot)$ and $f_B(\cdot)$ represent discretization rules for $\bm{A}$ and $\bm{B}$. It is suggested that $f_A(\cdot)$ followed the zero-order hold rule, while $f_B(\cdot)$ follows the Euler rule \cite{gu2023mamba}, expressed as:
\begin{equation}
	\begin{aligned}
		f_A({\Delta}_{t, d}, \bm{A}_d) &= \exp({\Delta}_{t, d} \cdot \bm{A}_d), \\
		f_B({\Delta}_{t, d}, \bm{B}_{t})&= {\Delta}_{t, d} \cdot \bm{B}_{t}.
	\end{aligned} \label{eq:dis}
\end{equation}

\subsection{Time Series Detrending}
A common time series model with trend and seasonality could be formulated as $\bm{x}_t = \bm{\tau}_t + \bm{s}_t + \bm{r}_t,\ t = 1, \cdots, T,$
%
where $\bm{x}_t$ denotes the observation at the $t$-th time-step, $\bm{\tau}_t$ is the trend component, $\bm{s}_t$ is the seasonal component with a period of $k$, and $\bm{r}_t$ represents the residual component.
In this paper, emphasis was placed on trend removal and HP trend filter was adopted as the detrending method, expressed as: 
\begin{equation}
	\hat{\bm{\tau}}_t = \mathop{\arg\min}\limits_{\bm{\tau}_t} \sum_{t=1}^{T} (\bm{x}_t - \bm{\tau}_t)^2 + \lambda \sum_{t=2}^{T-1}(\bm{\tau}_{t+1} - 2\bm{\tau}_t + \bm{\tau}_{t-1})^2, \label{eq:hp}
\end{equation}
where $\lambda$ is a hyper-parameter controlling the smoothness.

\section{Method}
\begin{figure}[ht]
	\vspace{-1.0em}
	\centering
	\includegraphics[width=9cm]{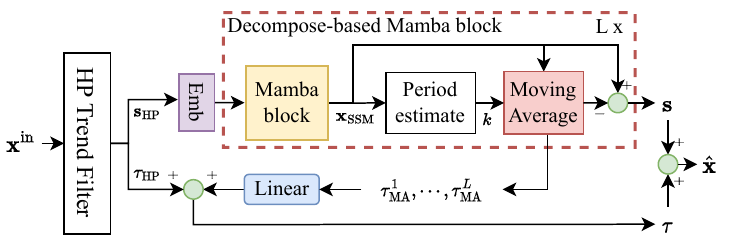}
	\caption{The structure of the proposed method includes a HP trend filter, multiple DMamba blocks, and an output module.}
	\vspace{-1.0em}
	\label{fig:framework}
\end{figure}

Fig. \ref{fig:framework} illustrates the overall framework of the proposed method, comprising a HP trend filter, multiple Decomposition-based Mamba (DMamba) blocks, and a final output module. Initially, input data was processed by the trend filter to extract initial trends and seasonality components. The initial seasonality was refined through DMamba blocks, while the initial trend was merged with trends from DMamba blocks. Finally, the fused trend and refined seasonality were summed to form the reconstructed output.

\subsection{HP Trend Filter}
The input data was initially segmented into subsequences using sliding windows of length $W$. {\color{black}Each input window $\bm{x}^{\rm in} \in \mathbb{R}^{W\times D}$ underwent detrending using HP trend filter. To ensure the extraction of global trends, the current input window was first combined with historical windows, and then the HP-based trend component $\bm{\tau}_{\rm HP}$ was computed via Eq. (\ref{eq:hp}). The HP-based seasonality was further derived as $\bm{s}_{\rm HP} = \bm{x}^{\rm in} - \bm{\tau}_{\rm HP}$.}

\subsection{DMamba Blocks}
\begin{figure}[ht]
	\vspace{-1.5em}
	\centering
	\includegraphics[width=8cm]{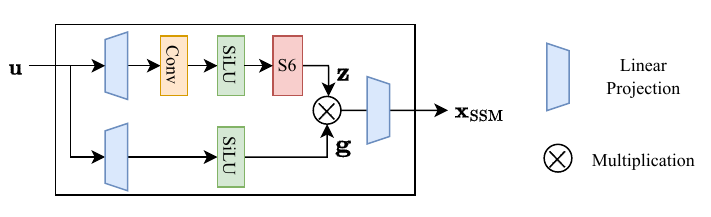}
		\vspace{-1.0em}
	\caption{The structure of Mamba block.}
	\label{fig:mamba}
	\vspace{-1.0em}
\end{figure}

The HP-based seasonality $\bm{s}_{\rm HP}$ was first projected to a $D_m$-dimensional space by an embedding network ${\rm Emb(\cdot)}$, and then it underwent $L$ DMamba blocks. Each block comprised a Mamba block \cite{gu2023mamba} and an Adaptive MA (AMA) module. The input of each block was the output of the previous block, denoted as $\bm{x}^{l} = \bm{y}^{l-1}$ with $\bm{y}^{0} = {\rm Emb}(\bm{s}_{\rm HP})$. For clarity, the superscript $l$ was omitted in the following content.
The input $\bm{x}$ first entered the Mamba block shown in Fig. \ref{fig:mamba} and could be formulated by Eq. (\ref{eq:mamba}):
\begin{equation}
	\begin{aligned}
		\bm{u},\ \bm{g} &= \sigma({\rm Conv}(\bm{x} \cdot \bm{W}_2 )), \ \sigma(\bm{x} \cdot \bm{W}_1), \\
		\bm{z}  &= {\rm S6}(\bm{u}), \\
		\bm{x}_{\rm SSM} &= (\bm{g} \odot \bm{z}) \cdot \bm{W}_3 , \label{eq:mamba}
	\end{aligned} 
\end{equation} 
where $\bm{W}_1, \bm{W}_2 \in \mathbb{R}^{D_m \times 2D_m}, \bm{W}_3 \in \mathbb{R}^{2D_m \times D_m}$ are trainable projection matrices, ${\rm Conv}(\cdot)$ represents a convolution layer with a kernel size of $4$,  $\sigma$ denotes the SiLU\cite{elfwing2018sigmoid} activation function, and ${\rm S6(\cdot)}$ denotes the S6 operation detailed in algorithm \ref{alg:s6}. The three parameters $\bm{\Delta}, \bm{B}, \bm{C}$ in Algorithm \ref{alg:s6} could be computed as:
\begin{align}
	\bm{\Delta} &= {\rm Softplus}(\bm{u}\cdot \bm{W}_\Delta) \label{eq:s6getd}, \\ 
	\bm{B} &= \bm{u}\cdot \mathbf{W}_B \label{eq:s6getb}, \\
	\bm{C} &= \bm{u}\cdot \mathbf{W}_C \label{eq:s6getc},
\end{align}
where ${\rm Softplus}(\cdot)$ denotes the Softplus activation function, 
$\bm{W}_B, \bm{W}_C \in \mathbb{R}^{2D_m \times N}$ and $\bm{W}_\Delta \in \mathbb{R}^{2D_m \times 2D_m}$ are trainable projection matrices.

\begin{algorithm}[!ht]
	\renewcommand{\algorithmicrequire}{\textbf{Input:}}
	\renewcommand{\algorithmicensure}{\textbf{Output:}}
	\caption{S6 operation}
	\label{alg:s6}
	\begin{algorithmic}[1]
		\REQUIRE $\bm{u}: (W, D)$ with $W$ time-steps and $D$ features ;
		\ENSURE $\bm{x}_{\rm SSM}: (W, D)$;
		
		\STATE $\bm{A}: (D, N) \leftarrow $ Parameter with $N$ states
		\STATE $\bm{\Delta}: (W, D) \leftarrow f_\Delta(\bm{u})$ using Eq. (\ref{eq:s6getd})
		\STATE $\bm{B}: (W, N) \leftarrow f_B(\bm{u})$ using Eq. (\ref{eq:s6getb}) 
		\STATE $\bm{C}: (W, N) \leftarrow f_C(\bm{u})$ using Eq. (\ref{eq:s6getc}) 
		\STATE $\overline{\bm{A}}, \overline{\bm{B}}: (W, D, N) \leftarrow {\rm discretize}(\bm{\Delta}, \bm{A}, \bm{B})$ using Eq. (\ref{eq:dis})
		
		\STATE $\bm{y} \leftarrow {\rm SSM}(\overline{\bm{A}}, \overline{\bm{B}}, \bm{C})(\bm{u})$ using Eq. (\ref{eq:s6})
		
		\STATE \textbf{return} $\bm{x}_{\rm SSM}$.
	\end{algorithmic}
\end{algorithm}

The AMA module then took $\bm{x}_{\rm SSM}$ as input and performed trend removal through average pooling:
\begin{equation}
	\bm{\tau}_{MA} = {\rm AvgPool}(\bm{x}_{\rm SSM}, k)
\end{equation}
with a kernel size of $k$. The kernel size is adaptively estimated by $k = \left\lfloor W/n \right\rfloor$, where $\left\lfloor \cdot \right\rfloor$ denotes the floor function, $n = \arg\max(\bm{x}^H_f \odot \bm{x}_f)$ indicates the peak position of the power density function and 
$\bm{x}_f$ is the Fourier transformation of $\bm{x}_{\rm SSM}$. The output of the current block is then obtained by $\bm{y} = \bm{x}_{\rm ssm} - \bm{\tau}_{\rm MA}$.

\subsection{Final Reconstructed Output}
The output of the final DMamba block served as the reconstructed seasonal component i.e., $\bm{s} = \bm{y}^L\cdot \bm{W}_s $, where $\bm{W}_s \in \mathbb{R}^{D_m\times D}$ is a trainable projection matrix. The final trend is a combination of all extracted trends:
\begin{equation}
	\bm{\tau} = \bm{\tau}_{\rm HP} + {\rm Conv}(\sum_{l=1}^{L} \bm{\tau}^l_{\rm MA}),
\end{equation}
where ${\rm Conv}(\cdot)$ is a CNN layer with an input channel of $D_m$, an output channel of $D$, and a kernel size of $3$.
In the training phase, the objective was to minimize the Mean Square Error (MSE) between the original input and the reconstructed output:
\begin{equation}
	\mathcal{L} = ||\bm{x}^{\rm in} - (\bm{s} + \bm{\tau})||^2_2.
\end{equation} 
During inference, the MSE serves as the anomaly score.

\section{Experiments}
\subsection{Setup}
\subsubsection{Datasets}
\begin{table}[ht]
	\caption{Datasets description.\label{tab:dataset}}
	\centering
	\begin{tabular}{ccccc}
		\toprule
		Dataset & Train & Test & Dimensions & Contamination (\%) \\
		\midrule
		NASA & $6,329$ & $18,743$ & $25/55$ & $9.52$ \\
		SMD & $128,267$ & $128,270$ & $38$ & $7.11$ \\
		SWaT & $6,840$ & $7,500$ & $25$ & $12.63$ \\
		\bottomrule
	\end{tabular}
\end{table}

\begin{table*}[t]
	\caption{
		Comparison results on three datasets. The best results were \textbf{bolded} while the
		second best were \underline{underlined}.
		\label{tab:main_res}}
	\centering
	\scalebox{1.0}{
		\begin{tabular}{l|ccc|ccc|ccc|ccc}
			\toprule
			\multicolumn{1}{c|}{\multirow{2}{*}{Method}} & \multicolumn{3}{c|}{NASA} & \multicolumn{3}{c|}{SWaT} & \multicolumn{3}{c|}{SMD} & \multicolumn{3}{c}{Avg.} \\
			& P-AF & R-AF & F1-AF & P-AF & R-AF & F1-AF & P-AF & R-AF & F1-AF & P-AF & R-AF & F1-AF \\
			\midrule
			LOF \cite{breunig2000lof} & $0.4020$ & $0.6313$ & $0.4886$ & $0.8450$ & $0.2855$ & $0.4268$ & $0.7002$ & $0.7376$ & $0.6959$ & $0.6491$  & $0.5515$ & $0.5371$\\
			CBLOF \cite{he2003discovering} & $0.4078$ & $0.6036$ & $0.4818$ & $0.6861$ & $0.0510$ & $0.0949$ & $0.8227$ & $0.7131$ & $0.6698$ & $0.6389$ & $0.4559$  & $0.4155$\\
			OCSVM \cite{scholkopf2001estimating} & $0.4735$ & $0.4739$ & $0.4341$ & $0.4558$ & $0.4316$ & $0.4434$ & $0.7923$ & $0.5003$ & $0.4900$ & $0.5739$  & $0.4686$  & $0.4558$\\
			IForest \cite{liu2008isolation} & $0.3811$ & $0.6353$ & $0.4744$ & $0.9977$ & $0.1424$ & $0.2492$ & $0.7905$ & $0.6589$ & $0.6359$ & $0.7231$  & $0.4789$  & $0.4532$\\
			HBOS \cite{goldstein2012histogram} & $0.5014$ & $0.9926$ & $0.6661$ & $0.9835$  & $0.1476$ & $0.2567$ & $0.7848$ & $0.5329$ & $0.5425$ &  $0.7566$ & $0.5577$  & $0.4884$\\
			LODA \cite{pevny2016loda} & $0.4736$ & $0.9551$ & $0.6326$ & $0.9777$ & $0.0263$ & $0.0513$ & $0.7379$ & $0.6064$ & $0.5674$ & $0.7297$  & $0.5293$  & $0.4171$\\
			ECOD \cite{li2022ecod} & $0.3806$ & $0.6238$ & $0.4720$ & $0.9940$ & $0.1122$ & $0.2016$ & $0.6358$ & $0.5551$ & $0.5349$ & $0.6701$  & $0.4304$ & $0.4028$\\
			Omnianomaly \cite{su2019robust} & $0.5074$ & $0.9999$ & $0.6732$ & $0.5085$ & $0.9738$ & $0.6681$ & $0.4957$ & $0.9221$ & $0.6438$ & $0.5039$  & $0.9653$ & $0.6617$\\
			FGANomaly \cite{du2021gan} & $0.5101$ & $1.000$ & $\underline{0.6756}$ & $0.5338$ & $0.9639$ & $0.6871$ & $0.6283$ & $0.9937$ & $0.7659$ & $0.5574$  & $0.9859$ & $\underline{0.7095}$\\
			Anomalytrans \cite{xu2021anomaly} & $0.5024$ & $0.9792$ & $0.6639$ & $0.5965$ & $0.5929$ & $0.5947$ & $0.5398$ & $0.9415$ & $0.6847$ & $0.5462$ & $0.8379$  & $0.6478$\\
			DCdetector \cite{yang2023dcdetector} & $0.4848$ & $0.9777$ & $0.6471$ & $0.4753$ & $0.5175$ & $0.4955$ &  $0.5019$ & $0.5367$ & $0.6412$ &  $0.4873$ & $0.6773$  & $0.5946$\\
			D3R \cite{wang2024drift} & $0.4334$ & $0.9776$ & $0.5868$ & $0.6241$ & $0.7970$ & $\underline{0.7000}$ & $0.7506$ & $0.9328$  & $\underline{0.8295}$ & $0.6027$  & $0.9025$ & $0.7054$\\
			\midrule
			Proposed method  & $0.5282$ & $0.9766$ & $\bm{0.6829}$ & $0.6441$ & $0.8641$ & $\bm{0.7380}$ & $0.8061$ & $0.9040$ & $\bm{0.8403}$ &  $0.6595$ & $0.9149$ & $\bm{0.7537}$\\
			\bottomrule
		\end{tabular}
	}
	\vspace{-1.5em}
\end{table*}

Experiments were conducted on three publicly available datasets. The \textbf{NASA} dataset \cite{hundman2018detecting}  contained sensor metrics recorded by Soil Moisture Active Passive satellite (SMAP) and Mars Science Laboratory rover (MSL). Three non-trivial subsets were utilized, namely \textit{A-4}, \textit{T-1}, and \textit{C-2}. as suggested in \cite{tuli2022tranad}.
The \textbf{Server Machine Dataset (SMD)} \cite{su2019robust} includes multiple server machine metrics from a large Internet company. Emphasis was placed on non-trivial subsets \textit{1-1}, \textit{1-6}, \textit{2-1}, \textit{3-2}, and \textit{3-7}, as suggested \cite{tuli2022tranad}.
Lastly, the \textbf{Secure Water Treatment (SWaT)} \cite{mathur2016swat} was a dataset that spans 11 days of continuous operation. In the experiments, SWaT was downsampled to the minute level and binary features were discarded. Detailed information is listed in Tab. \ref{tab:dataset}.

\subsubsection{Baselines}
The proposed method was compared to traditional and DNN-based methods. Traditional methods include LOF \cite{breunig2000lof}, CBLOF \cite{he2003discovering}, OCSVM \cite{scholkopf2001estimating}, IForest \cite{liu2008isolation}, HBOS \cite{goldstein2012histogram}, LODA \cite{pevny2016loda} and ECOD \cite{li2022ecod}.
DNN-based methods comprised two RNN-based methods: Omnianomaly \cite{su2019robust} and FGANomaly \cite{du2021gan}, two Transformer-based methods: AnomalyTrans \cite{xu2021anomaly} and DCdetector \cite{yang2023dcdetector}, and a time series decomposition-based method: D3R \cite{wang2024drift}.

\subsubsection{Implementation details}
All baselines were implemented based on PyOD \cite{zhao2019pyod} or their publicly available codes. We utilized Peak-Over-Threshold (POT) \cite{siffer2017anomaly} for threshold selection. The smoothness parameter $\lambda$ was set to $10^4$.
The block size $L$ was $3$, the model size $D_m$ was $32$, the state size was $16$ and the window size $W$ was $100$. The training was conducted for $7$ epochs with a batch size of $128$, using the AdamW \cite{loshchilov2017decoupled} optimizer with a learning rate of $10^{-4}$. 
{\color{black}Given that point-wise metrics can be influenced by lengthy anomalies and point-adjustment strategy \cite{su2019robust} could lead to misguided rankings \cite{kim2022towards}, the recently proposed affiliation-based precision (P-AF), recall (R-AF), and F1-score (F1-AF) \cite{huet2022local} were adopted as evaluation metrics. These metrics offer evaluations at the level of abnormal events rather than individual points, making them suitable for TSAD. Since F1-AF represents the harmonic mean of P-AF and R-AF, it was employed to evaluate the overall performance, a common practice in the TSAD community. }

\subsection{Results}
The proposed method was compared to $12$ baselines, as listed in Tab. \ref{tab:main_res}. Generally, sequence model-based methods outperformed traditional methods due to their ability to model temporal dependencies. The proposed method surpassed all compared methods on the three datasets, achieving a $6.2\%$ relative improvement in average F1-AF compared to the best baseline method. This demonstrated that the proposed S6 model-based detector was suitable for TSAD tasks and superior to previous SOTA detectors with RNN or Transformer backbones. Meanwhile, for the SWaT and SMD datasets containing significant trend variation, D3R and the proposed method achieved notable performance improvement by incorporating time series decomposition approaches compared to other methods. Overall, the experimental results showcased the robustness of the proposed method in non-stationary scenarios and its ability to detect complex time series anomalies. 

\begin{table}[ht]
	\vspace{-1em}
	\caption{Ablation study of the HP trend filter and AMA. The best results were \textbf{bolded} while the second best were \underline{underlined}.\label{tab:ablation}}
	\vspace{-1.5em}
	\begin{center}
		\scalebox{1.0}{
			\begin{tabular}{cc|c|c|c|c}
				\toprule
				\multicolumn{2}{c|}{Modules} & \multicolumn{4}{c}{F1-AF} \\
				HP Trend Filter & AMA  & \multirow{1}{*}{NASA} & \multirow{1}{*}{SWaT} & \multirow{1}{*}{SMD} & \multirow{1}{*}{Avg.} \\
				\midrule
				\XSolidBrush & \XSolidBrush & $0.6717$ & $0.6968$ & $0.7780$ & $0.7155$ \\
				\XSolidBrush & \Checkmark & $\underline{0.6777}$ & $0.6662$ & $0.7782$ & $0.7074$ \\
				\Checkmark & \XSolidBrush & $0.6700$ & $\underline{0.7302}$ & $\underline{0.8110}$ & $\underline{0.7371}$ \\
				\Checkmark & \Checkmark  & $\bm{0.6829}$ & $\bm{0.7380}$ & $\bm{0.8403}$ & $\bm{0.7537}$ \\
				\bottomrule
			\end{tabular}
		}
	\end{center}
	\vspace{-2.0em}
\end{table}

\begin{figure}[ht]
	\vspace{-0.5em}
	\centering
	\includegraphics[width=8.0cm]{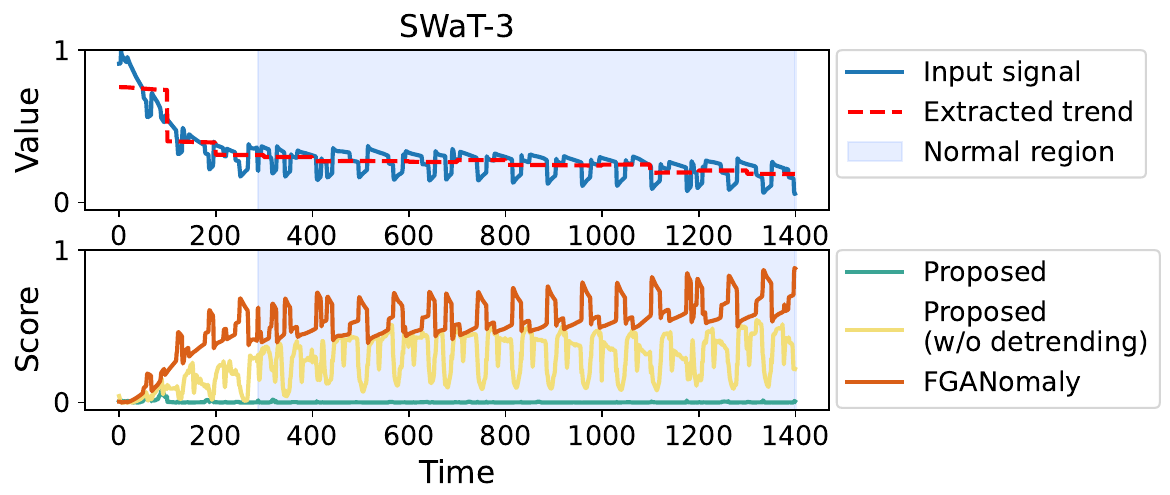}
	\vspace{-1.0em}
	\caption{{\color{black}A visualization from the SWaT dataset, which contains non-stationary data. The first row shows the input signal and extracted trend. The second row displays anomaly scores from various detectors. Baseline methods were influenced by trends, leading to false alarms, while our proposed method generated reliable scores by incorporating detrending mechanisms.}}
	\label{fig:cs}
	\vspace{-1.0em}
\end{figure}

To further verify the necessity of the HP trend filter and the AMA mechanism, an ablation analysis
was conducted and the results were shown in Tab. \ref{tab:ablation}. The results demonstrated that applying only one might not bring improvement. For instance, solely applying the HP trend filter in the NASA dataset or only utilizing AMA in the SWaT dataset led to a performance drop. However, when both were applied, the HP trend filter provided stable input for AMA's period estimation, where AMA and the detector could symbiotically optimize performance and capture more complex patterns, resulting in the best performance over all datasets. 
{\color{black}Additionally, a visualization in Fig. \ref{fig:cs} clearly showcased the effectiveness of the proposed detrending mechanisms.
}

\section{Conclusion}
This paper proposed solutions to effectively tackle two critical challenges in TSAD: modeling long-range dependency and generalizing to non-stationary data, which are bottlenecks for sequence models in TSAD. Our solutions involve leveraging a recently published S6 model to capture long-term context and introducing a novel multi-stage detrending mechanism to provide stable input for the sequence model. Experimental results underscored the suitability of the S6 model for TSAD and highlighted the importance of the proposed detrending mechanism. Furthermore, this study laid the foundation for developing more advanced S6 model-based detectors for TSAD.

\bibliographystyle{IEEEtran}
\bibliography{refs.bib}

\end{document}